\documentclass{article}

\usepackage[final,nonatbib]{nips}
\usepackage[square,numbers]{natbib}

\usepackage[utf8]{inputenc} % allow utf-8 input
\usepackage[T1]{fontenc}    % use 8-bit T1 fonts
\usepackage{hyperref}       % hyperlinks
\usepackage{url}            % simple URL typesetting
\usepackage{booktabs}       % professional-quality tables
\usepackage{amsfonts}       % blackboard math symbols
\usepackage{nicefrac}       % compact symbols for 1/2, etc.
\usepackage{microtype}      % microtypography

\usepackage{graphicx}
\usepackage{adjustbox}
\usepackage{colortbl} 
\usepackage{xcolor} 
\usepackage[skins]{tcolorbox}
\usepackage{amsmath}

\usepackage{amsthm}
\usepackage[utf8]{inputenc}
\usepackage[english]{babel}
\usepackage{amssymb}
\newtheorem{proposition}{Proposition}
\usepackage[ruled,vlined,noresetcount]{algorithm2e}

\title{GASL: Guided Attention for Sparsity Learning in Deep Neural Networks}

\author{
  Amirsina Torfi \\
  Department of Computer Science\\
  Virginia Tech\\
  Blacksburg, VA, USA \\
  \texttt{amirsina.torfi@gmail.com} \\
   \And
   Rouzbeh A.~Shirvani \\
   Howard University \\
   Washington DC, USA \\
   \texttt{rouzbeh.asghari@gmail.com}\\
   \AND
   Sobhan Soleymani \\
   West Virginia University \\
   Morgantown, WV, USA \\
   \texttt{ssoleyma@mix.wvu.edu} \\
   \And
   Naser M.~Nasrabadi \\
   West Virginia University \\
   Morgantown, WV, USA \\
   \texttt{nasser.nasrabadi@mail.wvu.edu} \\
}

\begin{document}

\maketitle

\begin{abstract}

The main goal of network pruning is imposing sparsity on the neural network by increasing the number of parameters with zero value in order to reduce the architecture size and the computational speedup. In most of the previous research works, sparsity is imposed stochastically without considering any prior knowledge of the weights distribution or other internal network characteristics. Enforcing too much sparsity may induce accuracy drop due to the fact that a lot of important elements might have been eliminated. In this paper, we propose \textit{Guided Attention for Sparsity Learning}~(GASL) to achieve \textbf{(1)} model compression by having less number of elements and speedup; \textbf{(2)} prevent the accuracy drop by supervising the sparsity operation via a guided attention mechanism and \textbf{(3)} introduce a generic mechanism that can be adapted for any type of architecture; Our work is aimed at providing a framework based on an interpretable attention mechanisms for imposing structured and non-structured sparsity in deep neural networks. For \textit{Cifar-100} experiments, we achieved the state-of-the-art sparsity level and 2.91$\times$ speedup with competitive accuracy compared to the best method. For \textit{MNIST} and \textit{LeNet} architecture we also achieved the highest sparsity and speedup level.

\end{abstract}

\section{Introduction}

Recent advances in deep neural networks came with ideas to train deep architectures that have led to near-human accuracy for image recognition, object categorization and a wide variety of other applications~\cite{lecun2015deep,maturana2015voxnet,schmidhuber2015deep,mnih2013playing,hinton2012deep}. One possible issue is that an over-parameterized network may make the architecture overcomplicated for the
task at hand and it might be prone to over-fitting as well. In addition to the model complexity, a huge amount of computational power is required to train such deep models due to having billions of weights. Moreover, even if a huge model is trained, it cannot be effectively employed for model evaluation on low-power devices mainly due to having exhaustive matrix multiplications~\cite{courbariaux2015binaryconnect}.

So far, a wide variety of approaches have been proposed for creating more compact models. Traditional methods include model compression \cite{ba2014deep,hinton2015distilling}, network pruning \cite{han2015learning}, sparsity-inducing regularizer~\cite{collins2014memory}, and low-rank approximation~\cite{jaderberg2014speeding,denton2014exploiting,ioannou2015training,tai2015convolutional}. The aforementioned methods usually induce random connection pruning which yields to few or no improvement in the computational cost. On the other hand, structured pruning methods proposed to compress the architecture with significant computational efficiency~\cite{wen2016learning,neklyudov2017structured}. 

One of the critical subjects of interest in sparsity learning is to maintain the accuracy level. In this paper, we discuss the \textit{intuitive reasons behind the accuracy drop and propose a method to prevent it}. The important step is to determine how the sparsity and accuracy are connected together in order to be able to propose a mechanism for controlling the sparsity to prevent severe accuracy drop. In order to connect the sparsity to accuracy, intuitively, the accuracy drop is caused by imposing too much sparsity on the network in a way that the remaining elements cannot transfer enough information for optimal feature extraction for the desired task. Another intuitive reasoning is to argue that the sparsity is not supervised with any attention towards the model performance during optimization. 

%To have a better intuition about the effective or ineffective weights, we consider the effective weights to be sufficiently larger ones com-
%pared to the other weights. This is reasonable because the weights are simple multipliers and in case of having larger values can affect the activations more than the other elements.

For effective network pruning and feature selection, different approaches such as employing the group lasso for sparse structure learning \cite{yuan2006model}, structure scale constraining \cite{liu2015sparse}, and structured regularizing deep architectures known as Structured Sparsity Learning (SSL) \cite{wen2016learning} have previously been proposed. For most of the previous research efforts, there is lack of addressing the direct effect of the proposed method on the combination of the sparsity and accuracy drop. One may claim that successful sparsity imposing with negligible accuracy drop might be
due to the initial over-parameterizing the network. Moreover, there is \textit{no control mechanism to supervise the sparsity operation connected to the model performance} which limits the available methods to intensive hyper-parameter tuning and multiple stages of training.

\textbf{Our contribution.} We designed and employed a supervised attention mechanism for sparsity learning which: \textbf{(1)} performs model compression for having less number of parameters \textbf{(2)} prevents the accuracy drop by sparsity supervision by paying an attention towards the network using variance regularization and \textbf{(3)} is a generic mechanism that is not restricted by the sparsity penalty or any other limiting assumption regarding the network architecture. To the best of our knowledge, \textit{this is the first research effort which proposes a supervised attention mechanism for sparsity learning}.

\textbf{Paper Organization.} At first, we provide a review of the related research efforts~(Section~\ref{sec:Related works}). Then, we introduce the attention mechanism which is aimed at forcing some sections of the network to be active~(Section~\ref{sec:Proposed attention mechanism}). Later in Section~\ref{sec:GASL: Guided Attention in Sparsity Learning}, we propose an algorithm only for the attention supervision. We complement our proposed method in Section~\ref{sec:Experimental results}, by providing experimental results for which we target the sparsity level, accuracy drop and robustness of the model to hyper-parameter tuning. As will be observed, the proposed mechanism prevents the severe accuracy drop in higher levels of sparsity. We will empirically show the robustness to exhaustive hyper-parameter tuning in addition to performance superiority of the proposed method in higher sparsity levels.

%\textbf{Notation.} The notation $|\zeta|$ is the dimensionality of the associated parameters~($\zeta$), and $\lambda$ considers to be a weighting coefficient for a regularizer.

\section{Related works}\label{sec:Related works}
\textbf{Network weights pruning.} Network compression for parameter reduction has been of great interest for a long time and a large number of research efforts are dedicated to it. In \cite{han2015learning,han2015deep,ullrich2017soft,molchanov2017variational}, network pruning has been performed with a significant reduction in parameters, although they suffer from computational inefficiency due to the mere weight pruning rather than the structure pruning.

\textbf{Network structure pruning.} In \cite{louizos2017bayesian,wen2016learning,neklyudov2017structured}, pruning the unimportant parts of the structure\footnote{This can be neurons in fully-connected layers or channels/filters in convolutional layers.} rather than simple weight pruning has been proposed and significant computational speedup has been achieved. However, for the aforementioned methods, the architecture must be fully trained at first and the potential training speedup regarding the sparsity enforcement cannot be attained. A solution for training speedup has been proposed by $\ell_0$-regularization technique by using online sparsification~\cite{louizos2017learning}. Training speedup is of great importance but adding a regularization for solely speeding up the training~(because of the concurrent network pruning) is not efficient due to adding a computational cost for imposing $\ell_0$-regularization itself. Instead, we will use an adaptive gradient clipping~\cite{pascanu2013difficulty} for training speedup.

\textbf{Attention.} In this paper, the goal is to impose the sparsity in an accurate and interpretable way using the attention mechanism. So far, attention-based deep architectures has been proposed for image~\cite{fu2017look,jia2015guiding,mnih2014recurrent,xu2015show} and speech domains~\cite{bahdanau2016end,chorowski2015attention,toro2005speech}, as well as machine translation~\cite{bahdanau2014neural,luong2015effective,vaswani2017attention}. Recently, the supervision of the attention mechanism became a subject of interest as well~\cite{liu2016neural,liu2017attention,chen2016guided,mi2016supervised} for which they proposed to supervise the attention using some external guidance. We propose the use of guided attention for enforcing the sparsity to map the sparsity distribution of the targeted elements\footnote{Elements on which the sparsity enforcement is desired: Weights, channels, neurons and etc.} to the desired target distribution.

\section{Proposed attention mechanism}\label{sec:Proposed attention mechanism}

%An example of a desired distribution for the weights is depicted in Fig.~\ref{fig:dist}.
%
%
%\begin{figure}[h]
%\begin{center}
%%\framebox[4.0in]{$\;$}
%\includegraphics[scale=0.15]{imgs/dist.png}
%%\fbox{\rule[-.5cm]{0cm}{4cm} \rule[-.5cm]{4cm}{0cm}}
%\end{center}
%\caption{An example for the weights sckewed distribution.}
%\label{fig:dist}
%\end{figure}

The main objective of the attention mechanism is to control and supervise the sparsity operation. For this aim, it is necessary to propose a method which is neither dependent on the architecture of the model nor to any layer type while maintaining the model accuracy and enforcing the compression objective. Considering the aforementioned goals, we propose the \textit{variance loss} as an auxiliary cost term to force the distribution of the weights\footnote{Or any elements on which we are enforcing sparsity.} to be skewed. A skewed distribution with a high variance and a concentration on zero~(to satisfy the sparsity objective) is desired. Our proposed scheme supervises the sparsity operation to keep a portion of the targeted elements~(such as weights) to be dominant~(with respect to their magnitude) as opposed to the other majority of the weights to simultaneously impose and control sparsity.~Intuitively, this mechanism may force a portion of weights to survive for sufficient information transmission through the architecture.

Assume enforcing the sparsity is desired on a parametric model; let's have the training samples with $\{x_i,y_i\}$ as pairs. We propose the following objective function which is aimed to create a sparse structure in addition to the \textit{variance regularization}:

\begin{align}\label{eq:gen-objective}
\begin{split}
&\Omega(\boldsymbol{\theta}) =  \frac{1}{N} \left (\sum_{i=1}^{N}\Gamma \left ( F(\boldsymbol{x_i};\boldsymbol{\theta}),\boldsymbol{y_i} \right ) \right ) +  R(\boldsymbol{\theta}) +  \lambda_{s} . \ominus(G(\boldsymbol{\theta})) + \lambda_{v} . \Psi^{-1}(H(\boldsymbol{\theta)}), \\
& \theta_{opt} = \underset{\boldsymbol{\theta}}{\mathrm{argmin}}\{\Omega(\boldsymbol{\theta})\},
\end{split}
\end{align}

in which $\Gamma(.)$ corresponds to the cross-entropy loss function and $\boldsymbol{\theta}$ can be any combination of the target parameters. Model function is defined as $F(.)$, $R(.)$ is some regularization function, $G(.)$ and $H(.)$ are some arbirtrary functions\footnote{Diffrentiable in general.} on parameters~(such as grouping parameters), $N$ is the number of samples, $\lambda$ parameters are the weighting coefficients for the associated losses and $\ominus(.)$ and $\Psi(.)$  are the sparsity and variance functions\footnote{The $\Psi(\boldsymbol{\theta})$ is simply taking the variance on the set of $\boldsymbol{\theta}$ parameters if H(.) is the identity function.}, respectively. The variance function is the utilized regularization term for any set of $\boldsymbol{\theta}$ parameters\footnote{Such as groups, weights and etc.}. The inverse operation on top of the $\Psi(.)$ in Eq.~\ref{eq:gen-objective} is necessary due to the fact that the higher variance is the desired objective. The power of the variance as a regularizer has been investigated in \cite{namkoong2017variance}. In this work, we expand the scope of variance regularization to the sparsity supervision as well.

%We call our method \textit{Attention-based Guided Sparsity Learning~(AGSL)}

\subsection{Model complexity}\label{sec:Model complexity}

Adding a new term to the loss function can increase the model complexity due to adding a new hyper-parameter~(the coefficient of the variance loss). For preventing this issue, we propose to have a dependent parameter as the variance loss coefficient. If the new hyperparameter is defined in terms of a variable dependent upon another hyperparameter, then it does not increase the model complexity. Considering the suggested approach, a dependency function must be defined over the hyperparameters definition.~The dependency is defined as $\lambda_{v} = f(\lambda_{s}) = \alpha \times \lambda_{s}$ in which $\alpha$ is a scalar multiplier.

\subsection{Structured attention for group lasso regularization}\label{sec:Structured attention}

\textbf{Group Sparsity.} Group sparsity has widely been utilized mostly for its feature selection ability by deactivating neurons\footnote{Or channels in convolutional layer.} via imposing sparsity on the whole cluster of weights in a group~\cite{yuan2006model,meier2008group}. Regarding the Eq.~\ref{eq:gen-objective}, the group sparsity objective function can be defined by follwoing expression:

\begin{align}\label{eq:variance-group}
\begin{split}
\ominus(G(\boldsymbol{w})) =  \sum_{l=1}^{N_l} \frac{1}{\sqrt{|G(W^{l})|}} \left (\sum_{j=1}^{M}\sqrt{\sum_{i=1}^{|w^{(j)}|}(w_{i}^{(j)})^{2}}\right )_l ,\\
\end{split}
\end{align}

in which $w^{(j)}$ is the $j_{th}$ group of weights in $w$ and $|w^{(j)}|$ is the number of weights in the associated group in the case of having M groups. The $l$ indicates the layer index, $|G(W^{l})|$ is a normalizer factor which is in essence the number of groups for the $l_{th}$ layer and $(.)_l$ demonstrates the elements~(weights) belonging to the the $l_{th}$ layer.

\textbf{Structured attention.} We propose a \textit{Structured Attention~(SA)} regularization, which adds the attention mechanism on group sparsity learning~(the sparsity imposition operation is smiliar to SSL~\cite{wen2016learning}). The attention is on the predefined groups. Under our general framework, it can be expressed by the following substitutions in Eq.~\ref{eq:gen-objective}:

\begin{align}\label{eq:variance-group}
\begin{split}
\Psi(H(\boldsymbol{w})) =  \sum_{l=1}^{N_l} \frac{1}{\sqrt{|G(W^{l})|}} \left ( \frac{1}{M}\sum_{j=1}^{M}\left (  \sqrt{\sum_{i=1}^{|w^{(j)}|}(w_{i}^{(j)})^{2}} - \frac{1}{M}\sum_{k=1}^{M}\sqrt{\sum_{i=1}^{|w^{(k)}|}(w_{i}^{(k)})^{2}}\right )^{2}\right )_l,
\end{split}
\end{align}

which is simply the variance of the group values for each layer, normalized by a factor and aggregated for all the layers.

\textbf{Generalizability.} It is worth noting that our proposed mechanism is not limited to the suggested structured method. It can operate on any $\ominus(.)$ function as sparsity objective because the definition of the attention is independent of the type of the sparsity. As an example, one can utilize an \textit{unstructured attention} which is simply putting the attention function $\Psi(.)$ on all the network weights without considering any special groups of weights or prior objectives such as pruning unimportant channels or filters.

\section{GASL: Guided Attention in Sparsity Learning}\label{sec:GASL: Guided Attention in Sparsity Learning}

The attention mechanism observes the areas of structure\footnote{Groups, weights or elements.} on which the sparsity is supposed to be enforced. we propose the \textit{Guided Attention in Sparsity Learning (GASL)} mechanism, which aims at the attention supervision toward mapping the distribution of the elements' values to a certain target distribution. The target distribution characteristics must be aligned with the attention objective function with the goal of increasing the variance of the elements for sparsity imposition.

\subsection{Increasing variance by additive random samples}\label{sec:additive-random-vector}

Assume we have the vector $V(\boldsymbol{\theta})=[\boldsymbol{\theta}_1,\boldsymbol{\theta}_2,...,\boldsymbol{\theta}_{|\boldsymbol{\theta}|}]^T$ that is the values of the elements in the group $[\boldsymbol{\theta}]=\{\boldsymbol{\theta}_1,\boldsymbol{\theta}_2,...,\boldsymbol{\theta}_{|\boldsymbol{\theta}|}\}$ and for which we want to maximize the variance. In \cite{paisley2012variational}, variational Bayesian inference has been employed for the gradient computation of variational lower bound. Inspired by \cite{wang2013variance}, in which random vectors are used for stochastic gradient optimization, we utilize the additive random vectors for variance regularization. The random vector is defined as $V^r(\boldsymbol{\theta})=[V^r_1,V^r_2,...,V^r_{|\boldsymbol{\theta}|}]^T$.The formulation is as below:

\begin{equation}\label{eq:variance-add}
\hat{V}(\boldsymbol{\theta}) = V(\boldsymbol{\theta}) + M. \left(V^r(\boldsymbol{\theta})-E(V^r(\boldsymbol{\theta}))\right),
\end{equation}

where $M$ is a $|\boldsymbol{\theta}| \times |\boldsymbol{\theta}|$ matrix. The resulted vector $\hat{V}(\boldsymbol{\theta})$ does not make any changes in the mean of the parameters distribution since it has the same mean as the initial $V(\boldsymbol{\theta})$ vector. 

\begin{proposition}
Assume the variance maximization of the $\hat{V}(\boldsymbol{\theta})$ is desired. The choice of matrix M, does not enforce any upperbound on the variance vector $\hat{V}(\boldsymbol{\theta})$.

\textbf{Proof.} For that aim, the task breaks to the subtask of finding the optimal $M$ for which the trace of the $\hat{V}(\boldsymbol{\theta})$ is maximized. For the mini-batch optimization problem, we prove that the proposed model is robust to the selection of $M$. The maximizer $M$ can be obtained when the trace of the variance matrix is maximized and it can be demonstrated as follows:

\begin{align}\label{eq:variance-max}
\begin{split}
M_{*} = & \underset{M}{\mathrm{argmax}} \left \{ Tr\left ( Var[\hat{V}(\boldsymbol{\theta})] \right ) \right \}\\
= & \underset{M}{\mathrm{argmax}} \left \{ Tr\left (  Cov[\hat{V}(\boldsymbol{\theta}),\hat{V}(\boldsymbol{\theta})] \right ) \right \}\\
= & \underset{M}{\mathrm{argmax}} \left \{ Tr \left ( Var[V(\boldsymbol{\theta})] + \Upsilon + \Upsilon^T + M.Var[V^r(\boldsymbol{\theta})].M^T  \right ) \right \}\\
, & \Upsilon = M. Cov[V^r(\boldsymbol{\theta}),V(\boldsymbol{\theta})].
\end{split}
\end{align}

As can be observed from Eq.~\ref{eq:variance-max}, as long as $M$ is a positive definite matrix, the additive random can add to the value of the matrix trace without any upper bound. The detailed mathematical proof is available in the Appendix.$\blacktriangle$

\end{proposition}

Considering the mathematical proof, one can infer that the mere utilization of the variance loss term in Eq.~\ref{eq:gen-objective}, as the attention mechanism and without the additive random vector, can supervise the sparsity operation. However, we will empirically show that the additive random vectors can improve the accuracy due to the supervision of the attention. \textit{The supervision of the variance is important regarding the fact that the high variance of the parameters may decrease the algorithm speed and performance for sparsity enforcement. This is due to the large number of operations that is necessary for gradient descent to find the trade-off between the sparsity and variance regularization}. From now one, without the loss of generality, we assume $M$ to be \textit{identity matrix} in our experiments.\\

\textbf{The choice of random vector $V^r$.} In practice, the choice of $V^r$ should be highly correlated with $V$. Furthermore, Eq.~\ref{eq:variance-max} shows that without being correlated, the terms associated with $Cov[V^r(\boldsymbol{\theta}),V(\boldsymbol{\theta})]$ may go to zero which affects the high variance objective, negatively. The algorithm for random vector selection is declared in Algorithm.~\ref{algorithm:random-vector-selection}.\textit{ The distribution $pdf(.)$ should be specified regarding the desired output distribution}. We chose \textit{log-normal distribution} due to its special characteristics which create a concentration around zero and a skewed tail~\cite{limpert2001log}. If the variance of the random vector $V^r(\boldsymbol{\theta})$ is less than the main vector $V(\boldsymbol{\theta})$, no additive operation will be performed. In case the $[\boldsymbol{\theta}]$ parameters variance is high-enough compared to the $V(\boldsymbol{\theta})$ vector, then there is no need for additive random samples. This preventing mechanism has been added due to the practical speedup.

\begin{algorithm}\label{algorithm:random-vector-selection} 
 \KwData{Extract mini-batch;}
 \textbf{Parameters group}: Form $V(\boldsymbol{\theta})=[\boldsymbol{\theta}_1,\boldsymbol{\theta}_2,...,\boldsymbol{\theta}_{|\boldsymbol{\theta}|}]^T$\;
 \textbf{Random sampling}: Draw $\beta_{k}$ $\thicksim$ $pdf(\beta)$ for $k$ in $\{1,2,...,\boldsymbol{\theta}\}$ \;
 \textbf{Random vector creation}: Form $V^r(\boldsymbol{\theta})=[\beta_1,\beta_2,...,\beta_{|\boldsymbol{\theta}|}].V(\boldsymbol{\theta})$ \;
 %\While{$Var[V^r(\boldsymbol{\theta})] > Var[V(\boldsymbol{\theta})]$}{
  %evaluate the current impostor pair output distance: $imp\_dis$\;
  \eIf{$Var[V^r(\boldsymbol{\theta})] > Var[V(\boldsymbol{\theta})]$}{
   \textbf{Replacement vector}: Calculate $\hat{V}(\boldsymbol{\theta}) = V(\boldsymbol{\theta}) + M. \left(V^r(\boldsymbol{\theta})-E(V^r(\boldsymbol{\theta}))\right)$\;
  \textbf{Replacement operation}: Replace $\hat{V}(\boldsymbol{\theta})$ with $V(\boldsymbol{\theta})$\;
  \textbf{Return}: $\hat{V}(\boldsymbol{\theta})$\;
   }{
   \textbf{Return}: $V(\boldsymbol{\theta})$\;
  }
  \textbf{Computation}: Update gradiant\;
 %}
 
 \caption{GASL algorithm. }
\end{algorithm}

\subsection{Deacreasing ambiguity by the variance inverse}

In practice, we do not want to add ambiguity to the system. In another word, ideally, the new loss term should not add any extra information for decision making. This problem is closely related to the Fisher information in statistics which is the information that an observed random variable $X$ has about the unknown parameter $\boldsymbol{\theta}$ \cite{rissanen1996fisher,efron1978assessing,frieden2004science}. The parameter $\boldsymbol{\theta}$ can be the weights or groups of weights in our model. In Eq.~\ref{eq:gen-objective}, we chose the variance inverse for this aim.

\begin{proposition}

The choice of the variance inverse with the objective of having a high variance, has inverse proportional relation with the Fisher information,i.e., having more variance, decreases the Fisher information added by the variance loss term.

\textbf{Proof.} Assume $\mathcal{O}$ is used for classification purposes and has the hidden variable of $[\boldsymbol{\theta}]=\{\boldsymbol{\theta}_1,\boldsymbol{\theta}_2,...,\boldsymbol{\theta}_n\}$. Let the likelihood distribution be $p(\mathcal{O}|\boldsymbol{\theta})$. The Fisher information is equal to the second moment of the log-likelihood distrubution as follows:

\begin{align}\label{eq:fisher-information}
I(\boldsymbol{\theta}) = E_\mathcal{O}\left [ \left ( \frac{\partial^2 }{\partial \boldsymbol{\boldsymbol{\theta}}^2} log\{p(\mathcal{O}|\theta)\} \right )^2|\boldsymbol{\theta} \right ].
\end{align}

Assume we have an unbiased estimator $\hat{\boldsymbol{\theta}}(\mathcal{O})$ which has the condition of $E\left [ \hat{\boldsymbol{\theta}}(\mathcal{O}) - \boldsymbol{\theta} | \boldsymbol{\theta} \right ]$. Regarding that, we have the following inequality using the Cramér-Rao Bound \cite{frieden2004science}:

\begin{align}\label{eq:fisher-bound}
Var[\hat{\boldsymbol{\theta}}]\geq \frac{1}{I(\boldsymbol{\theta})}.
\end{align}

The Eq.\ref{eq:fisher-bound} expresses the fact that if the variance of an estimation is more than a value $\mathcal{V}$, then the Fisher information lower bound will be $\mathcal{V}^{-1}$.$\blacktriangle$

\end{proposition}

We have shown in Section \ref{sec:additive-random-vector} that the unbiased estimate can be made by simply generating a random vector with the same mean value which satisfies the assumed condition on $\hat{\boldsymbol{\theta}}(\mathcal{O})$. The Eq.\ref{eq:fisher-bound} does not necessarily put an upper bound on the Fisher information but decreasing the lower bound is still justified with the \textit{objective of not increasing the ambiguity}.

\begin{figure}[ht]
\begin{center}
%\framebox[4.0in]{$\;$}
\includegraphics[scale=0.32]{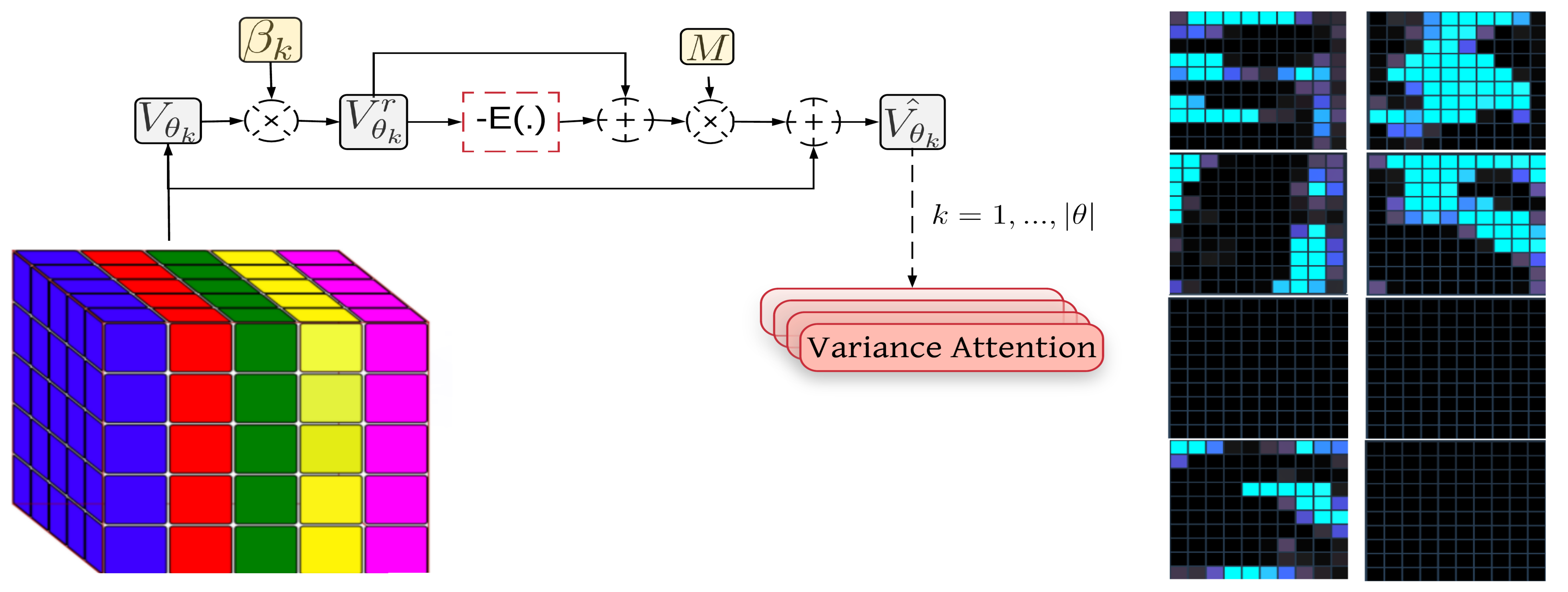}
%\fbox{\rule[-.5cm]{0cm}{4cm} \rule[-.5cm]{4cm}{0cm}}
\end{center}
\caption{The combination of GASL and structured attention. The cube demonstrates the output feature map of a convolutional layer. The weights associated with each channel, form a group. For visualization, the right column is the activation visualization of the attention-based sparsity enforcement on output channels and the left one is the results of imposing sparsity without attention. As can be observed, some of the channels are turned off and the remaining ones are intesified.}
\label{fig:SA_GASL}
\end{figure}

\textbf{Combination of GASL and SA.} GASL algorithm can operate on the top of the structured attention for attention supervision. The schematic is depicted in Fig.~\ref{fig:SA_GASL}. Furthermore, a visualized example of the output channels from the second convolutional layer in the MNIST experiments has also demonstrated in Fig.~\ref{fig:SA_GASL}. The structured attention is dedicated to the output channels of a convolutional layer.

% \Upsilon
% Cov[V_r({\theta}),V({\theta})]
%One possible approach is to simply impose the variance loss on all the weights such as the aforementioned descriptions. This is similar to general $\ell_{1}-$ and $\ell_{2}-regularization$ with different objective and functionality. The proposed additional term is supposed to control sparsity harshness by forcing a portion of elements to be alive.

%The parameters $\boldsymbol{\theta}$

%The controller function can operate on any set of predefined values such as groups, weights and etc. One possible approach is to simply impose the variance loss on all the weights such as the aforementioned descriptions. This is similar to general $\ell_{1}-$ and $\ell_{2}-regularization$ with different objective and functionality. The proposed additional term is supposed to control sparsity harshness by forcing a portion of elements\footnote{Groups, weights and etc.} to be alive.

\section{Experimental results}\label{sec:Experimental results}

We use three databases for the evaluation of our proposed method: MNIST~\cite{lecun1998gradient}, CIFAR-10 and CIFAR-100~\cite{krizhevsky2009learning}. For increasing the convergence speed without degrading the overall performance, we used gradient clipping~\cite{pascanu2013difficulty}. A common approach is to clip individual gradients to some fixed predefined range $[-\zeta,\zeta]$. As the learning rate becomes smaller continuously, the effective gradient\footnote{Which is $gradient \times learning\_rate$} will approach zero and training convergence may become extremely slow. For tackling this issue, we used the method proposed in \cite{kim2016accurate} for gradient clipping which defined the range dynamically as $[-\zeta/\gamma,\zeta/\gamma]$ for which $\gamma$ is the current learning rate. We chose $\zeta=0.1$  in our experiments. Hyper-parameters are selected by cross-validation. For all our experiments, \textit{the output channels of convolutional layers and neurons in fully connected layers are considered as groups.}

\subsection{MNIST dataset}

For experiments on MNIST dataset, we use $\ell_2$-regularization with the default hyperparameters. Two network architectures have been employed: LeNet-5-Caffe\footnote{https://github.com/BVLC/caffe/blob/master/examples/mnist} and a multilayer perceptron~(MLP). For the MLP network, the group sparsity is enforced on each neuron's outputs for feature selection; Same can be applied on the neurons' inputs as well. The results are shown in Table.~\ref{table:MNIST-resutls}. The percentage of sparsity is reported layer-wise. One important observation is the superiority of the \textit{SA} method to the \textit{SSL} in terms of accuracy, while the sparsity objective function for both is identical and the only difference is the addition of structural attention~(Eq.~\ref{eq:variance-group}). As a comparison to \cite{louizos2017bayesian}, we achieved closed sparsity level with better accuracy for the MLP network. For the LeNet network, we obtained the highest sparsity level with competitive accuracy compared to the best method proposed in \cite{molchanov2017variational}.

\begin{table}[htbp]
  \centering
  \small\addtolength{\tabcolsep}{-1pt}
  \caption{Experiments on LeNet-5-Caffe architecture with 20-50-800-500 number of output filters and hidden layers and MLP with the architecture of 784–500–300 as the number of hidden units for each layer. The sparsity level is reported layer-wise.}
  %The last layer for both architectures have been dropped for demonstration as those are only used for classification and no sparsity has been performed on them.}
\begin{tabular}{*{11}{c}}
    \toprule
    & \multicolumn{2}{c}{LeNet} & \multicolumn{2}{c}{MLP}  \\
    \cmidrule(lr){2-3}
    \cmidrule(lr){4-5}
    %\cmidrule(lr){8-9}
    Method & Error($\%$) & Sparsity($\%$) / speedup & Error($\%$) & Sparsity($\%$) / speedup \\
    \midrule
    baseline & 0.8 & 0-0-0-0 / 1.00$\times$ & 1.54 & 0-0-0 / 1.00$\times$\\
    $\ell_1$-regularization & 2.44 & 15-31-37-53 / 1.07$\times$ & 3.26 & 21-23-15 / 1.01$\times$\\
    Network Pruning~\cite{han2015learning} & 1.21 & 61-58-80-67 / 1.17$\times$ & 1.71 & 20-32-69 / 1.03$\times$\\
    Bayesian Compression~\cite{louizos2017bayesian} & 0.9 & 60-74-89-97 / 2.31$\times$ & 1.8 & \textbf{71-81-94} / 1.18$\times$\\
    Structured BP~\cite{neklyudov2017structured} & 0.86 & 85-62-64-43 / 2.03$\times$ & 1.55 & 68-77-89 / \textbf{1.28}$\times$\\
    Structured Sparsity Learning~\cite{wen2016learning} & 1.00 & 71-58-61-34 / 1.83$\times$ & 1.49 & 52-61-74 / 1.12$\times$\\
    Sparse Variational Dropout~\cite{molchanov2017variational} & \textbf{0.75} & 66-36-59-75 / 1.43$\times$ & 1.57 & 31-56-57 / 1.05$\times$\\
    $\ell_0$-regularization~\cite{louizos2017learning} & 1.02 & 8-62-96-17 / 1.31$\times$ & \textbf{1.41} & 32-34-37 / 1.07$\times$\\
    \bottomrule
    Structured Attention & 1.05 & 78-62-72-50 / 1.91$\times$ & 1.56 & 22-31-62 / 1.04$\times$\\
    Structured Attention + GASL & 0.92  & \textbf{76-88-86-95} / \textbf{2.41}$\times$ & 1.53 & 64-80-95 / 1.23$\times$\\
    \bottomrule
  \end{tabular}
  \label{table:MNIST-resutls}
\end{table}

\subsection{Experiments on Cifar-10 and Cifar-100}

For experiments in this section, we used VGG-16~\cite{simonyan2014very} as the baseline architecture. Random cropping, horizontal flipping, and per-image standardization have been performed for data augmentation in the training phase and in the evaluation stage, only center cropping has been used~\cite{krizhevsky2012imagenet}. Batch-normalization has also been utilized after each convolutional layer and before the activation~\cite{ioffe2015batch}. The initial learning rate of $0.1$ has been chosen and the learning rate is dropped by a factor of 10 when the error plateaus for 5 consecutive epochs. As can be observed from Table.~\ref{table:CIFAR-resutls}, the combination of the GASL algorithm and SA dominates regarding the achieved sparsity level and demonstrates competitive results in terms of accuracy for Cifar-100. We terminate the training after 300 epochs or if the averaged error is not improving for 20 consecutive epochs, whichever comes earlier. For Cifar-10, we obtained the second best results for both accuracy and sparsity level.
\begin{table}[htbp]
  \centering
  \small\addtolength{\tabcolsep}{-1pt}
  \caption{Experiments on Cifar-10 and Cofar-100 using VGG-16 network.}
\begin{tabular}{*{11}{c}}
    \toprule
    & \multicolumn{2}{c}{Cifar-10} & \multicolumn{2}{c}{Cifar-100}  \\
    \cmidrule(lr){2-3}
    \cmidrule(lr){4-5}
    %\cmidrule(lr){8-9}
    Method & Error~($\%$) & Sparsity~($\%$) / speedup & Error~($\%$) & Sparsity~($\%$) / speedup\\
    \midrule
    baseline & 8.75 & 0 / 1.00$\times$ & 27.41 & 0 / 1.00$\times$\\
    $\ell_1$-regularization & 11.43 & 22 / 1.36$\times$ & 31.75 & 17 / 1.35$\times$\\
    Network Pruning~\cite{han2015learning} & 10.76 & 32 / 1.54$\times$ & 28.46 & 22 / 1.46$\times$\\\
    Bayesian Compression~\cite{louizos2017bayesian} & 8.42 & \textbf{82} / 3.16$\times$ & 25.72 & 41 / 2.02$\times$\\
    Structured BP~\cite{neklyudov2017structured} & 8.62 & 46 / 2.86$\times$ & 25.47 & 29 / 2.45$\times$\\
    Structured Sparsity Learning~\cite{wen2016learning} & 9.12 & 74 / \textbf{3.31}$\times$ & 26.42 & 46 / 2.43$\times$\\
    Sparse Variational Dropout~\cite{molchanov2017variational} & \textbf{7.79} & 61 / 2.14$\times$ & \textbf{24.91} & 42 / 2.11$\times$\\
    $\ell_0$-regularization~\cite{louizos2017learning} & 8.83 & 52 / 2.41$\times$ & 26.73 & 38 / 1.92$\times$\\
    \bottomrule
    Structured Attention & 8.62 & 48 / 2.80$\times$ & 25.76 & 32 / 2.37$\times$\\
    Structured Attention + GASL & 8.32 & 76 / 3.06$\times$ & 25.41 & \textbf{48} / \textbf{2.91}$\times$\\
    \bottomrule
  \end{tabular}
  \label{table:CIFAR-resutls}
\end{table}

\textbf{The advantage of the proposed method for higher sparsity levels.} For Cifar-100 experiments, we continued the process of enforcing sparsity for achieving the desired level of compression\footnote{The increasing of the $\lambda_s$ might have been necessary.}. We chose three discrete level of sparsity and for any of which, the accuracy drop for different methods is reported. Table.~\ref{table:accuracy-drop} demonstrates the comparison of different methods with regard to their accuracy drops at different levels of sparsity. For some levels of sparsity, it was observed that some methods performed better than the baseline. We deliberately selected higher levels of sparsity for having some performance drop as opposed to the baseline for all the implemented methods. As can be observed, our method shows its performance superiority in accuracy for the higher levels of sparsity. \textit{In another word, the proposed method outperforms in preventing the accuracy drop in the situation of having high sparsity level. 
}
%\begin{figure}[htbp]
%\begin{center}
%%\framebox[4.0in]{$\;$}
%\includegraphics[scale=0.25]{imgs/accuracy-drop.png}
%%\fbox{\rule[-.5cm]{0cm}{4cm} \rule[-.5cm]{4cm}{0cm}}
%\end{center}
%\caption{The accuracy drop, compared to the baseline, for the different levels of sparsity for Cifar-100.}
%\label{fig:accuracy-drop}
%\end{figure}

\begin{table}[htbp]
  \centering
  \small\addtolength{\tabcolsep}{-1pt}
  \caption{The accuracy drop, compared to the baseline, for different levels of sparsity using Cifar-100. Our method here is SA + GASL. The accuracy drop is measured compared to the baseline results with no sparsity.}
\begin{tabular}{*{11}{c}}
    \toprule
    & \multicolumn{8}{c}{Method}  \\
    \cmidrule(lr){2-9}
    %\cmidrule(lr){8-9}
     Sparsity & $\ell_1$ & \cite{han2015learning} & \cite{louizos2017bayesian} & \cite{neklyudov2017structured} & \cite{wen2016learning} & \cite{molchanov2017variational} & \cite{louizos2017learning} & Ours\\
    \midrule
    60\% & 1.51 & 1.31 & 0.63 & 0.67 & 0.74 & \textbf{0.59} & 0.61 & 0.71\\
    70\% & 3.46 & 2.73 & 1.73 & 1.79 & 1.89 & 1.81 & 1.74 & \textbf{1.68}\\
    80\% & 4.21 & 4.78 & 2.47 & 2.68 & 2.64 & 2.73 & 2.81 & \textbf{2.23}\\
    \bottomrule
  \end{tabular}
  \label{table:accuracy-drop}
\end{table}

\textbf{Robustness to the hyperparameter tuning.} Regarding the discussion in Section~\ref{sec:Model complexity}, it is worth to investigate the effect of $\lambda_{v}$ on the accuracy drop. \textit{In another word, we investigate the relative importance of tuning the variance loss coefficient}. The accuracy drop is reported for Cifar-100 experiments using different $\alpha$ values and sparsity levels. The results depicted in Table.~\ref{table:robustness-alpha}, empirically shows the robustness of the proposed method to the selection of $\alpha$, as the dependent factor, for which in the dynamic range of $[0.1,10]$, the accuracy drop is not changing drastically. \textit{This clearly demonstrates the robustness of the proposed method to the selection of the new hyperparameter associated with the attention mechanism as it is only a dependent factor to the sparsity penalty coefficient}.

\begin{table}[htbp]
  \centering
  \small\addtolength{\tabcolsep}{-1pt}
  \caption{Experiments on Cifar-100 for investigating the robustness of the proposed method to the hyperparameter selection.}
\begin{tabular}{*{11}{c}}
    \toprule
    & \multicolumn{5}{c}{$\alpha=\lambda_v/\lambda_s$}  \\
    \cmidrule(lr){2-6}
    %\cmidrule(lr){8-9}
     Sparsity & 0.01 & 0.1 & 1.0 & 10 & 100\\
    \midrule
    60\% & 2.1 & 0.86 & 0.71 & 0.67 & 1.61\\
    70\% & 3.46 & 1.91 & 1.68 & 1.79 & 3.21\\
    80\% & 4.21 & 2.51 & 2.13 & 2.09 & 4.41\\
    \bottomrule
  \end{tabular}
  \label{table:robustness-alpha}
\end{table}

\section{Conclusion}

In this paer, we proposed a guided attention mechanism for controlled sparsity enforcement by keeping a portion of the targeted elements to be alive. The GASL algorithm has been utilized on top of the structured attention for attention supervision to prune unimportant channels and neurons of the convolutional and fully-connected layers. We demonstrated the superiority of the method for preventing the accuracy drop in high levels of sparsity. Moreover, it has been shown that regardless of adding a new term to the loss function objective, the model complexity remains the same and the proposed approach is relatively robust to exhaustive hyper-parameter selection. Without the loss of generality, the method can be adapted to any layer type and different sparsity objectives such as weight pruning for unstructured sparsity or channel, neuron or filter cancellation for structured sparsity.

\section{Acknowledgement}

This work was supported by the Center for Identification Technology Research (CITeR), a National Science Foundation~(NSF) Industry/University Cooperative Research Center (I/UCRC).

\bibliographystyle{unsrtnat}
\bibliography{ref}

\newpage
\section*{Appendix}

\subsection*{Proof of Proposition 1}

\begin{align}\label{apdx:eq:1}
\begin{split}
Var[\hat{V}(\boldsymbol{\theta})] = & Cov[\hat{V}(\boldsymbol{\theta}),\hat{V}(\boldsymbol{\theta})] \overset{a}{=} E[(\hat{V}(\boldsymbol{\theta})-E[\hat{V}(\boldsymbol{\theta})])(\hat{V}(\boldsymbol{\theta})-E[\hat{V}(\boldsymbol{\theta})])^T]\\
 \overset{b}{=} & E[\{V(\boldsymbol{\theta})-E[V(\boldsymbol{\theta})] + M. \left(V^r(\boldsymbol{\theta})-E[V^r(\boldsymbol{\theta})]\right)\}\{V(\boldsymbol{\theta})-E[V(\boldsymbol{\theta})] + M. \left(V^r(\boldsymbol{\theta})-E[V^r(\boldsymbol{\theta})]\right)\}^T]\\
\overset{c}{=} & E[(\overline{V(\boldsymbol{\theta})} + M. \overline{V^r(\boldsymbol{\theta})})(\overline{V(\boldsymbol{\theta})} + M. \overline{V^r(\boldsymbol{\theta})})^T]\\
 \overset{d}{=} & Var[V(\boldsymbol{\theta})] + M.Var[V^r(\boldsymbol{\theta})].M^T + M. Cov[V^r(\boldsymbol{\theta}),V(\boldsymbol{\theta})] + Cov[V(\boldsymbol{\theta}),V^r(\boldsymbol{\theta})].M^T\\
 \overset{e}{=} & Var[V(\boldsymbol{\theta})] + M.Var[V^r(\boldsymbol{\theta})].M^T + M. Cov[V^r(\boldsymbol{\theta}),V(\boldsymbol{\theta})] + Cov^T[V^r(\boldsymbol{\theta}),V(\boldsymbol{\theta})].M^T\\
=&Var[V(\boldsymbol{\theta})] + M.Var[V^r(\boldsymbol{\theta})].M^T + \zeta + \zeta^T,\zeta = M. Cov[V^r(\boldsymbol{\theta}),V(\boldsymbol{\theta})]
\end{split}
\end{align}

In the above, (a) comes directly from the covariance definition which is $Cov(\boldsymbol{X},\boldsymbol{Y}) = E[(\boldsymbol{X}-E[\boldsymbol{X}])(\boldsymbol{Y}-E[\boldsymbol{Y}])^T]$. The equality (b) is resulted from the expansion of $\hat{\boldsymbol{V}}(\theta)$. In (c), the bar sign for a variable means: $\overline{\boldsymbol{X}} = \boldsymbol{X} - E[\boldsymbol{X}]$. The line (d) is concluded using the following expressions:

\begin{align}
&E[\overline{\boldsymbol{X}}][\overline{\boldsymbol{Y}}^T] = Cov[\boldsymbol{X},\boldsymbol{Y}]\\
&E[\overline{\boldsymbol{X}}][\overline{\boldsymbol{X}}^T] = Cov[\boldsymbol{X},\boldsymbol{X}] = Var(\boldsymbol{X})\\
&E[A\boldsymbol{X}] = AE[\boldsymbol{X}]\\
&Cov[\boldsymbol{X},\boldsymbol{Y}]=Cov^T[\boldsymbol{Y},\boldsymbol{X}]
\end{align}

\subsection*{Log-Normal Distribution}

A random variable X with log-normal distribution and the mean $\mu$ and standard deviation of $\sigma$, X can be written as follows:
\begin{align}
X = e^{\{\mu + Y \sigma \}}
\end{align}

in which, $Y$ has a normal distribution. The reason behind choosing this distribution is to force the desired distribution of the parameters to be skewed with a concentration around zero for having the sparsity condition in addition to forcing a small portion of the parameters to be active. A log-normal distribution with $\mu=0.1$ and $\sigma=1$, can be demonstrated as below:

\begin{figure}[ht]
\begin{center}
%\framebox[4.0in]{$\;$}
\includegraphics[scale=0.3]{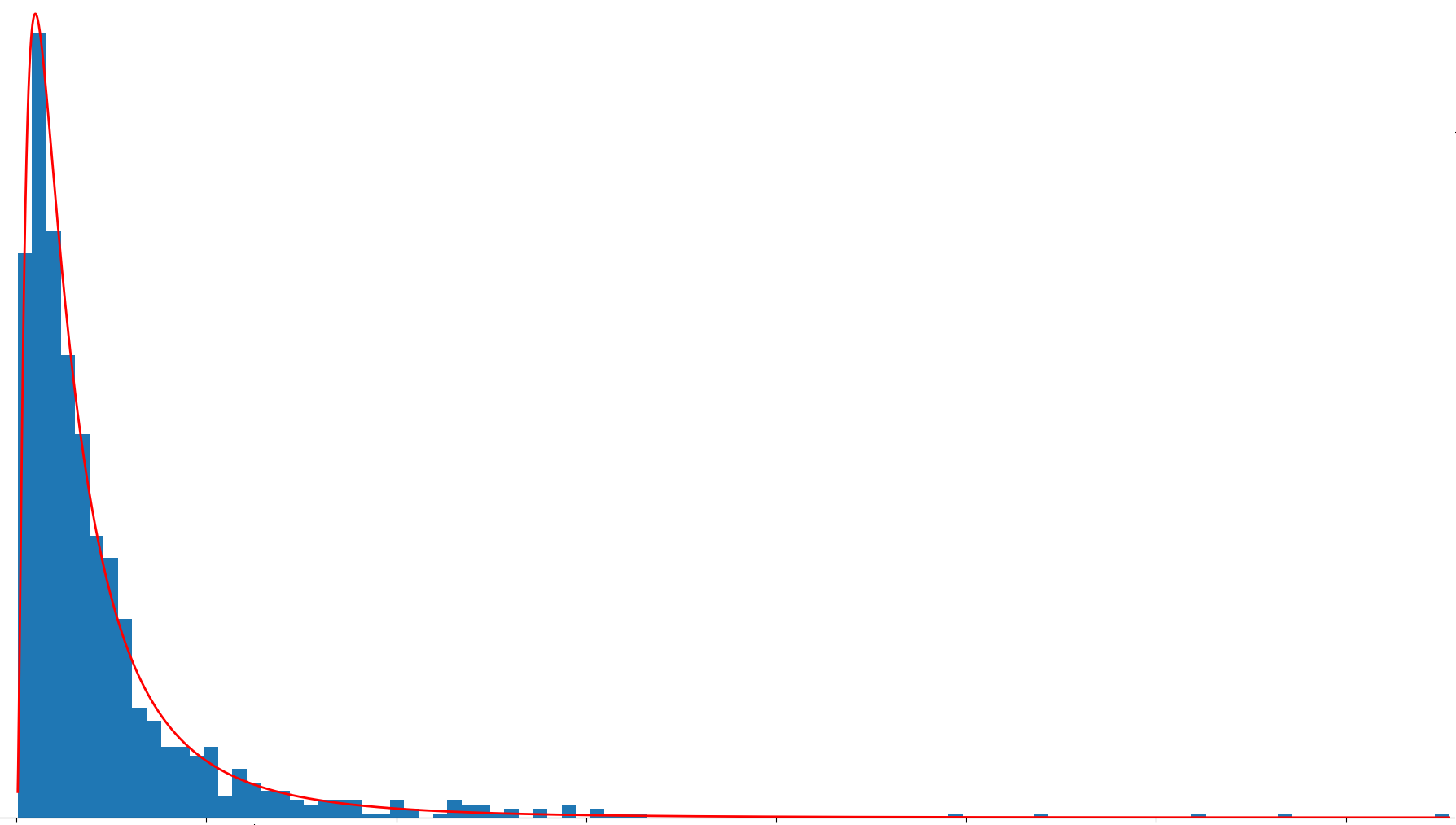}
%\fbox{\rule[-.5cm]{0cm}{4cm} \rule[-.5cm]{4cm}{0cm}}
\end{center}
\caption{An example for the an skewed distribution.}
\label{fig:dist}
\end{figure}

\end{document}